\documentclass[sigconf]{acmart}

\usepackage[T1]{fontenc}
\usepackage{graphicx}
\usepackage{booktabs}
\usepackage[misc]{ifsym}

\usepackage{orcidlink}
\usepackage{mathtools} 
\usepackage{siunitx} 
\usepackage{tikz} 

\usepackage{array}
\usepackage{amsmath}
\usepackage{amsthm}
\usepackage{amssymb}
\usepackage{subfigure}
\usepackage{xspace}
\usepackage{paralist}
\usepackage{caption,latexsym}
\usepackage[ruled,vlined,oldcommands]{algorithm2e}
\usepackage{times}
\usepackage{url}

\author{Monika Shah}
\authornote{University of Memphis}
\email{mshah2@memphis.edu}

\author{Somdeb Sarkhel}
\authornote{Adobe Research}
\email{sarkhel@adobe.com}

\author{Deepak Venugopal}
\authornote{University of Memphis}
\email{dvngopal@memphis.edu}

\begin{document}
\title{On Explaining Visual Captioning with Hybrid Markov Logic Networks}

\begin{abstract}

Deep Neural Networks (DNNs) have made tremendous progress in multimodal tasks such as image captioning. However, explaining/interpreting how these models integrate visual information, language information and knowledge representation to generate meaningful captions remains a challenging problem. Standard metrics to measure performance typically rely on comparing generated captions with human-written ones that may not provide a user with a deep insights into this integration. In this work, we develop a novel explanation framework that is easily interpretable based on Hybrid Markov Logic Networks (HMLNs) - a language that can combine symbolic rules with real-valued functions - where we hypothesize how relevant examples from the training data could have influenced the generation of the observed caption. To do this, we learn a HMLN distribution over the training instances and infer the shift in distributions over these instances when we condition on the generated sample which allows us to quantify which examples may have been a source of richer information to generate the observed caption.
Our experiments on captions generated for several state-of-the-art captioning models using Amazon Mechanical Turk illustrate the interpretability of our explanations, and allow us to compare these models along the dimension of explainability.

\end{abstract}


\maketitle

\section{Introduction}

The performance of deep neural networks (DNNs) on multimodal tasks such as visual captioning and visual question answering has improved tremendously over the past decade. However, explaining to an end-user {\em how these models learn} to integrate visual and language understanding remains challenging. Such explanations can have far-reaching implications in improving the usability of these models in real-world domains that rely on interpretability, e.g. health informatics, education, law, etc. Indeed, explainable AI (XAI) is regarded as a necessary step to drive AI adoption in critical domains. For instance, as is well-known, the European GDPR regulations requires transparency of data processing tools which makes XAI mandatory in such applications.

While it is well-known that interpreting DNN representations is notoriously difficult, symbolic AI has a rich history of representing interpretable knowledge. {\em Neuro-Symbolic AI} which has been regarded by some as the third-wave in AI~\cite{garcez2023neurosymbolic} combines DNN representations with symbolic AI. In this work, we develop a related model to generate {\em example-based explanations} for visual captioning. In social sciences, there is a rich body of work that strongly suggests that explanations to humans are more effective when they use examples within explanations~\cite{miller2019explanation}. Inspired by this, here, we use Hybrid Markov Logic Networks (HMLNs), a language that integrates relational symbolic representations with continuous functions (that can be derived from other sub-symbolic representations) to explain how a model could have learned to generate a caption by citing specific examples it has seen in the training data.

\begin{figure}
    \centering
    \includegraphics[scale=0.28]{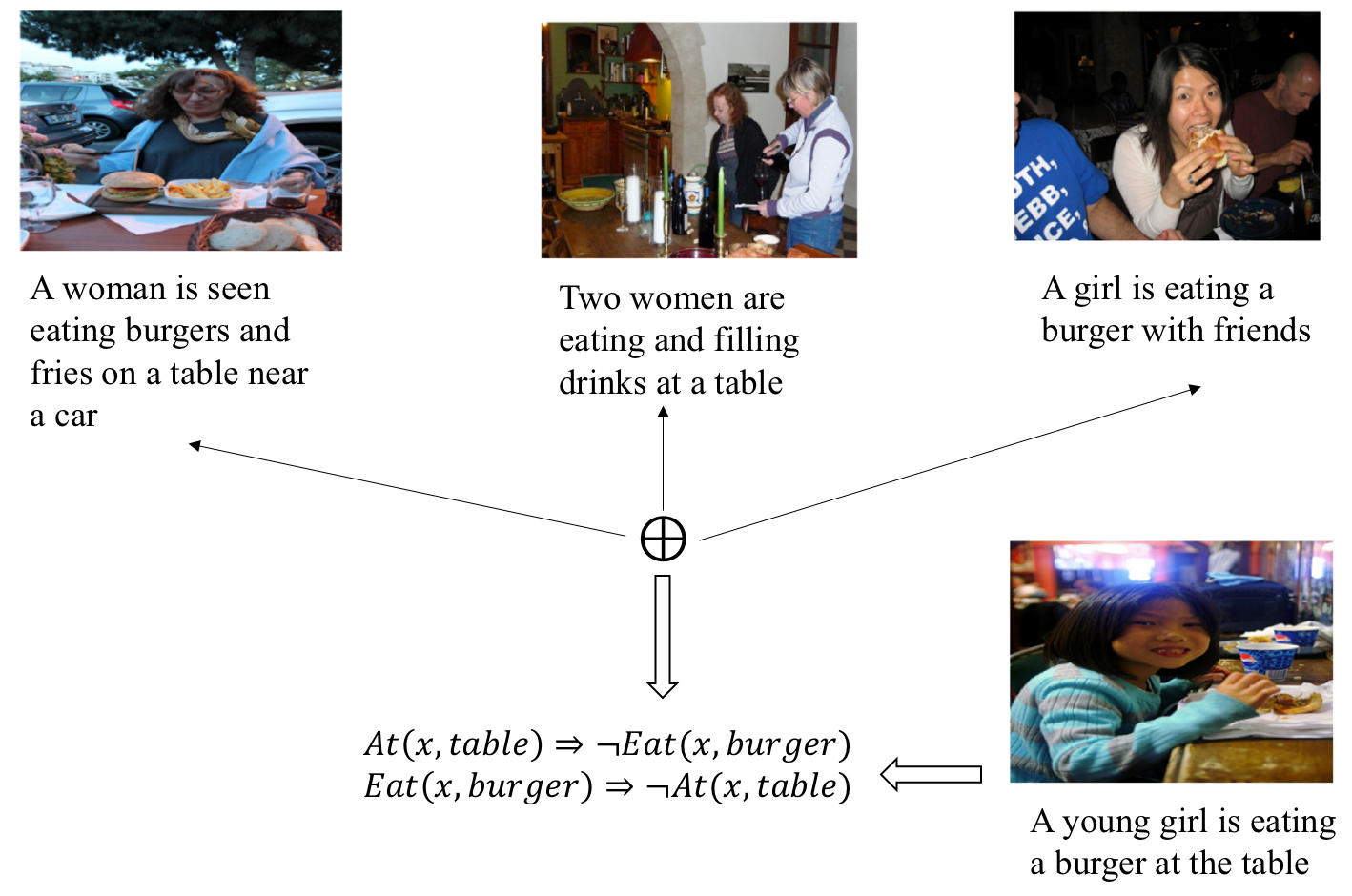}
    \caption{Illustration of our approach. The top row contains training data relevant to the test image shown at the bottom. The bias introduced by the test instance on the distribution over the training data is quantified using a probabilistic knowledge-base specified as a HMLN. Contrasting examples are chosen to explain the test caption. In the above example, the first and third images explain the relations in the test caption while the second image may not explain this.}
    \label{fig:example1}
\end{figure}

The main idea behind our approach is illustrated in Fig.~\ref{fig:example1}. Given a generated caption, we quantify shifts in the prior distribution (i.e., adding of bias) over the training data (where the distribution is represented as a HMLN) when conditioned on the generated caption. To hypothesize how the model could have learned to generate the caption, we select  examples with contrasting bias as an explanation. Further, to explain using the integration between the visual features of the image and text, we combine symbolic properties extracted from the text with real-valued functions that relate the symbolic properties to the visual features in the image using CLIP~\cite{radford2021learning} embeddings. The semantics of HMLNs allow us to encode these real-valued terms within symbolic properties and we parameterize these hybrid properties through max-likelihood estimation. 

To generate explanations, we weight the training images using importance weights that quantify bias. That is, we condition the prior distribution on (virtual) evidence observed from the test instance. We then generate samples from the conditioned distributions and weight it against the prior distribution. Using this, we select training examples that positively (and negatively) explain the generated caption based on the difference between the marginal estimates computed from the conditioned distribution and the estimates computed from the importance weighted samples.

We perform a comprehensive user study using Amazon Mechanical Turk to demonstrate the interpretability of explanations generated by our model. We show that our model is interpretable by both technical and non-technical users. Further, using this study, we compare the explainability of 4 state-of-the-art caption generating models on the MSCOCO dataset.




\subsection{Related Work}

There are several related approaches that try to deeply analyze multimodal systems such as visual captioning beyond comparison with ground-truth generated by humans. Specifically, reference-free metrics try to correlate the visual content with the generated caption to analyze its quality. Arguably, one of the most popular reference-free metrics is CLIPScore~\cite{hessel2021clipscore} based on the well-known CLIP model. Specifically, this uses CLIP to measure coherence between the visual representation of image and the textual representation of the caption.
Similarly, VIFIDEL~\cite{madhyastha2019vifidel}, measures the visual fidelity by comparing a representation of the generated caption with the visual content without the need of reference caption for evaluation.
Another stream of work relies on quantifying out of distribution (ood) instances to better understand the model's learning. 
In~\cite{shalev2022baseline} visually grounded tokens is used to detect out-of-distribution images in image captioning models. In particular, several manifestations of out of distribution instances such as those containing novel objects,  corrupted or anomalous images, low-quality images, etc. can be identified with this approach.
Related to explainability, the general field of XAI has made a lot of progress with techniques such as LIME~\cite{ribeiro2016should}, SHAP~\cite{lundberg2017unified} and integrated gradients~\cite{sundararajan2017axiomatic}. Typically these approaches work with standard classification problems. A related approach Grad-CAM~\cite{selvaraju2017grad} visually explains decisions by displaying relevant regions in the image. In generative AI, there has been some work in quantifying bias in generated samples~\cite{grover2019bias}.
In the context of multimodal tasks, there are fewer explanation methods. For Visual Question Answering (VQA) systems, explanation methods that have been developed include \cite{li2018tell}. The approach in ~\cite{wu2018faithful}, jointly highlights text words in the explanation of an answer along with image regions similar to Grad-CAM. However, our approach is distinct to this in the sense that instead of explaining how the answer (in our case caption) was generated by looking at the test instance only, we relate this to hypothesize how the model could have learned the generation from examples in the training data.

\section{Background}

\subsection{Hybrid Markov Logic Networks}
A HMLN is a set of pairs $(f_i,w_i)$, where $f_i$ is a formula that can contain both symbolic as well as numeric terms and $w_i$ is a weight associated with that formula to represent its uncertainty. If $f_i$ contains both symbolic and numeric terms, we call it a hybrid formula. Given a domain of constants that can be substituted for variables in the formulas, we can {\em ground} a formula by substituting all its variables with constants from their respective domains. A predicate symbol specifies a relation over terms. A {\em ground predicate} is one where all the variables in its arguments have been replaced by constants/objects from their respective domains. In the HMLN distribution, a ground predicate is considered as a random variable. Ground predicates can be symbolic (0/1) or numeric. We use the terms ground predicate or ground atom interchangeably. The {\em value of a ground formula} is equal to 0/1 if it contains no numeric terms while a ground formula with numeric terms has a numeric value. We can represent the distribution of a HMLN as a product of potential functions, where each potential function specifies a ground formula. Specifically, the distribution takes a log-linear form where each grounding has a value equal to $e^{w\times s}$, where $w$ is the weight of the formula that the potential represents and $s$ is the value of the formula (for a purely symbolic formula this is 0/1). We can specify the product over all potentials in the distribution as follows.
\begin{equation}
    P({\bf X}={\bf x})=\frac{1}{Z}\exp\left(\sum_iw_is_i(x)\right)
\end{equation}
where ${\bf x}$ is a {\em world}, i.e., an assignment to the variables defined in the HMLN distribution, $s_i({\bf x})$ is the value of the $i$-th formula given the world ${\bf x}$, $Z$ is the normalization constant also called as the partition function which is intractable to compute since it is a summation over all possible worlds.



\subsection{Gibbs Sampling}

Gibbs sampling~\cite{casella1992explaining} is one of the most widely used Markov chain Monte Carlo (MCMC) algorithms. In Gibbs sampling, we sample one variable in the distribution at a time given assignments to all other variables in the distribution.
Specifically, given a set of $n$ non-observed variables $X_1$ $\ldots$ $X_n$, the Gibbs sampling algorithm begins with a random assignment ${\bf \overline{x}}^{(0)}$ to them. We then sample a randomly selected atom say $X_i$ from the conditional distribution of $X_i$ given assignments to all other variables. This is given by


$$P(X_i|\overline{\bf x}_{-X_i}^{(t)}) =\frac{1}{Z_{X_i}}\exp\left(\sum_f w_fN_f(\overline{\bf x}_{-X_i}^{(t)}\cup X_i)\right)$$

where $\overline{\bf x}_{-X_i}^{(t)}$ represents an assignment to all variables except $X_i$, $Z_{X_i}$ is the normalization constant. The sample at iteration $t+1$ is $\overline{\bf x}_{-X_i}^{(t)}$ combined with the sampled assignment to variable $X_i$.

 
Typically, the sampler is allowed to run for some time (called the {\em burnin} time) to allow it to {\em mix} which ensures that it forgets its initialization and starts to collect samples from the target distribution. Once this happens, the sampler is said have {\em mixed}. The samples from a mixed Gibbs sampler can be used to estimate any expectation over the target distribution. A commonly used formulation is to write the marginal distribution of a variable (or set of variables) as an expectation over the indicator functions for that variable (or set of variables). We can show that as we estimate the expectations with more samples, the expectations converge to their true values.

\subsection{Importance Weighting}

Importance sampling re-weights samples drawn from a distribution (called the proposal) that is different from the target distribution. Specifically, suppose $P({\bf X})$ represents the target joint distribution over ${\bf X}$ and $Q({\bf X})$ is a proposal distribution, the sample ${\bf x}$ drawn from $Q(\cdot)$ has a weight equal to $\frac{P({\bf x})}{Q({\bf x})}$. Clearly, if the proposal distribution is closer to the target distribution, then the weight approaches 1. Just like in Gibbs sampling, we can estimate expectations of functions w.r.t. the target distribution. Here, we sum over the function values in the samples weighting each sample by its importance weights and normalize by the sum of importance weights.

\section{Approach}

\subsection{Explanation Templates}

We define templates of hybrid formulas (with symbolic and continuous terms) to define the HMLN structure. Note that the HMLN framework is flexible enough to represent arbitrary FOL structures. However, in our case, we limit the HMLN to explainable structures and therefore, similar to the approach in ~\cite{wang2008hybrid}, we pre-define templates and then slot-fill the templates from training data.
\begin{definition}
\label{def:1}
    A conjunctive property ($\mathcal{C}$) is a hybrid formula of the form, $f_1(X_i,\ldots X_k)$ $*$ $(X_1\wedge X_2\ldots X_k)$, where $X_i$ and $X_{i+1}$ are ground predicates that share at least one common term, $f_1()$ is a real-valued function over them.
\end{definition}

\begin{definition}
\label{def:2}
    An explanation property ($\mathcal{I}$) is defined as the XOR over a pair of ground predicates, i.e., $f_2(X_i,X_j)$ $*$ $(X_i\wedge \neg X_j)$ $\vee$ $(X_j\wedge \neg X_i)$ where $X_i,X_j$ share at least one variable between them and $f_2()$ is a real-valued function over them.
\end{definition}
The $\mathcal{I}$ property encodes our reasoning that observing a predicate explains away the existence of the other predicate. Specifically, we can write the symbolic part of the property as $(X_i\Rightarrow \neg X_j)$ $\wedge$ $(X_j\Rightarrow \neg X_i)$ if we ignore the case where the head of the implications are false. The (soft) equality over the functions is a continuous approximation of one predicate explaining away the other predicate.

The training data $\mathcal{D}$ consists of multiple captions (written by humans) for each image. For each image in the training data, we select one of the captions and extract ground predicates for this captions using a textual scene graph parser~\cite{schuster2015generating}. 
We slot-fill the $\mathcal{C}$ template by chaining predicates with shared objects in the same image. We limit the size of the chain to a maximum of two predicates since beyond this, the chains seem to lack coherence. For each $\mathcal{C}$ property, we also fill its corresponding $\mathcal{I}$ template. Thus, a pair of predicates with shared objects extracted from a training instance can be written as a pairwise potential function in the HMLN distribution, where the potential values are determined by the learned parameters and the values of their $\mathcal{C}$, $\mathcal{I}$ properties. An example for potential functions is illustrated in Fig.~\ref{fig:virt}.

\subsection{Learning Explainable Potentials}

We parameterize the HMLN structure using max-likelihood estimation. Typically, to learn parameters of the HMLN (or other statistical relational models), we would learn the parameters by maximizing the likelihood over the full training data. In other words, we ground the full HMLN to compute all the parameters. However, in our case this is not desirable due to two reasons. First, this will add irrelevant groundings making it infeasible to learn the parameters. Second, a single parameterization could be limiting when we want to generate fine-grained explanations specific to a generated caption. Therefore, we instead develop an {\em inference guided} parameterization where we learn parameters that are relevant to an inference query. Some related approaches such as ~\cite{chechetka2011query}, have explored similar learning methods. Here, we {\em normalize} the HMLN structure w.r.t the test instance that is being explained. In the normal form, we make a closed world assumption where given a test image and the generated caption, only the groundings that contain objects that are identified in the image are relevant to the test image and the other groundings by default have a value equal to 0 (i.e., we can ignore these groundings). To do this, we identify and localize objects in the image using Faster R-CNN~\cite{ren2015faster} and learn query-specific parameters for groundings that contain these objects. 

\begin{figure}
    \centering
    \includegraphics[scale=0.27]{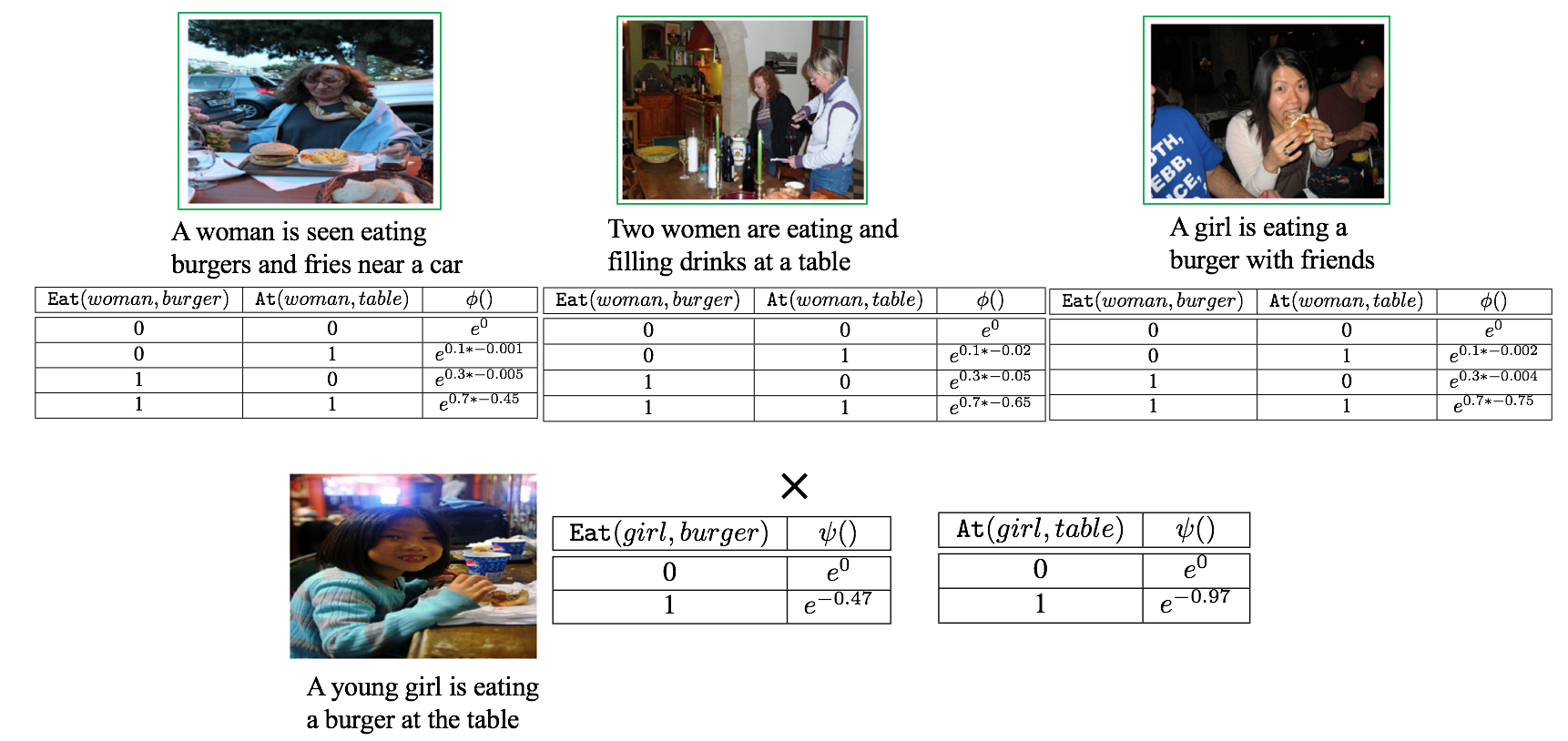}
    \caption{\label{fig:virt} Three training instances are shown in the top row. Each of these instances has a corresponding pairwise potential function shown below them. The potential function combines a weight along with the value of the properties encoded in the function. The test image has virtual evidence in the form of a univariate potential function. We multiply the virtual evidence with the training data potentials to obtain the evidence-instantiated potentials.}
    \label{fig:b}
\end{figure}

\noindent{\bf Parameterization.} One approach to parameterize the potentials is to learn a different weight for each potential. However, we want all the potentials that encode the same symbolic formulas to have the same weight. This is illustrated in Fig.~\ref{fig:b}. This {\em weight sharing} among the potential functions allows us to tie together all the functions that represent the same symbolic concepts but are grounded over different instances in the training data. To learn the weights, we maximize the likelihood over an observed world, ${\bf X}={\bf x}\in\{0,1\}^n$ which is an assignment to the predicates in every potential function. Note that the observed world assigns a 0/1 value to the variables in the potential function. To compute this value, we average over the captions underlying the potential. Specifically, if a relation is mentioned in all captions corresponding to the potential, we assign it a 1 else a 0. We now learn the weights of the potential functions by maximizing the following log-likelihood objective.
\begin{align}
\label{eq:likel}
    \ell({\bf w},{\bf x})=\log\frac{1}{Z_{\bf w}}\prod_i\phi_i({\bf x}) \nonumber\\= \sum_fw_f*\left(\sum_{\phi_j\sim f}\mathcal{I}_j({\bf x})+\mathcal{C}_j({\bf x})\right) - \log Z_{\bf w}
\end{align}
where ${\bf w}$ is the set of weights, $Z_{\bf w}$ is the normalization constant, $f$ is the symbolic formula associated with weight $w_f$ and $\phi_j\sim f$ denotes that the potential $\phi_j$ shares the symbolic formula $f$, i.e., $\phi_j$ is a grounding of $f$. Note that $\forall$ $\phi_j\sim f$, $w_f$ is a shared weight. $\mathcal{I}_j({\bf x})$, $\mathcal{C}_j({\bf x})$ are the $\mathcal{I}$, $\mathcal{C}$ values in $\phi_j$.

\noindent{\bf Real-Valued Terms.} We define the real-valued functions in definitions 2.1, 2.2 based on the visual features of the image in the training data that corresponds to the potential function. Specifically, suppose ground predicates $X_{1f},X_{2f}$ occur in formula $f$, we relate the symbolic representation with the visual representation in the training data corresponding to the potential function of $f$. To do this, we use CLIP~\cite{radford2021learning} to compute neural embeddings for the image in the training data as well as the text describing $X_{1f},X_{2f}$. We convert these neural embeddings into real-valued terms for the HMLN. Let $g_j(X_1)$, $g_j(X_2)$ be the cosine distance between the symbolic and visual embeddings for $X_1$ and $X_2$ respectively. We express the $\mathcal{I}$, $\mathcal{C}$ values as follows.
$$
\mathcal{I}_j({\bf x})=
\begin{cases}
-(g_j(X_1)$- $g_j(X_2))^2, & \text{if $f$ is true in {\bf x}}\\
0, & \text{if $f$ is false in {\bf x}}\\
\end{cases}
$$

The $\mathcal{I}$ value implies that the potential function has a value equal to the following expression.
\begin{equation}
\exp{(-w_f*(g_j(X_1)-g_j(X_2))^2)} = 
\exp{\left(\frac{-(g_j(X_1)-g_j(X_2))^2}{(2*(\sqrt{1/2w_f})^2}\right)}    
\end{equation}
Thus, the $\mathcal{I}$ value is a Gaussian penalty with standard deviation $\sqrt{1/2w_f}$. Larger differences between $g_j(X_1)$ and $g_j(X_2)$ imposes a greater penalty on the $\mathcal{I}$ property. This agrees with the symbolic relationship specified in the $\mathcal{I}$ property which is the negation of one ground predicate explaining the occurrence of another.

$$
\mathcal{C}_j({\bf x})=
\begin{cases}
\min\{\log\sigma(\epsilon-g_j(X_1)),\\\log\sigma(\epsilon-g_j(X_2))\}, & \text{if $f$ is true in {\bf x}}\\
0, & \text{if $f$ is false in {\bf x}}\\
\end{cases}
$$

The $\mathcal{C}$ value represents a threshold based on a constant $\epsilon$, i.e.,  $\log\sigma(g_j(X_1)-\epsilon)$ $=$ $-log(1+e^{a*(\epsilon-g(X))})$, where $a$ is the softness of the sigmoid function (in our experiments we set $\epsilon$ to 0.7 since this indicated that the text was a good explanation of the image). Thus, the $\mathcal{C}_j({\bf x})$ value quantifies the extent to which the symbolic concepts in $f$ describes the visual details in the image. The potential function value is equal to $\min$ $\{e^{\log\sigma(\epsilon-g_j(X_1)},e^{\log\sigma(\epsilon-g_j(X_2)}\}$ $=$ $\min\{\sigma(\epsilon-g_j(X_1)),\sigma(\epsilon-g_j(X_2))\}$.

Note that Lukasiewicz approximations for continuous logic~\cite{bach2017hinge} extends Boolean logic to the continuous case where each variable is allowed to take a continuous value in the interval $[0,1]$.The Lukasiewicz t-norm and t-co-norm define $\wedge$ and $\vee$ operators for continuous logic. The $\mathcal{C}$ and $\mathcal{I}$ values can be considered as variants of this approximation. Specifically, for the $\mathcal{C}$ value, we consider the minimum value (instead of the sum) since we want to penalize a formula even if one of its predicates is a poor descriptor for the image. Similarly, for the $\mathcal{I}$ value, we use the Gaussian penalty instead of a simple difference between the values.

\noindent{\bf Contrastive Divergence.} We can maximize $\log \ell({\bf w},{\bf x})$ using a standard gradient ascent method. The gradient given the weights ${\bf w}$ is as follows.

\begin{align}
\label{eq:grad}
   \frac{\partial(\ell({\bf w},{\bf x}))}{\partial{w_i}} = (\sum_i\mathcal{I}_i({\bf x})+\mathcal{C}_i({\bf x}))  - \mathbb{E}_{{\bf w}}[\mathcal{I}_i(\sum_i{\bf x})+\mathcal{C}_i({\bf x})]
\end{align}

As seen in the equation Eq.~\eqref{eq:grad}, to compute the gradient exactly, we need to compute the difference between the $\mathcal{I}$ and $\mathcal{C}$ property values in the data and their expected values given the current weights ${\bf w}$. However, the expectation is infeasible to compute exactly (a $\#P$-hard problem in the worst case) since we essentially need to sum over all possible worlds. Therefore, we instead estimate the expectation approximately. There are two ways to approximate the expectation. One is using MAP inference, i.e., we estimate the expected values from a single point, namely the mode of the distribution. This approach is known as the Voted Perceptron~\cite{singla2005discriminative}. However, a more robust estimator is to compute the expectation based on an average over worlds sampled from the distribution given weights ${\bf w}$. Typically a small number of MCMC steps can be used to obtain the approximate gradient direction. This approach is known as Contrastive Divergence~\cite{hinton2002training} and is known to yield higher quality weight estimates~\cite{pedro2009markov}. Therefore, we apply contrastive divergence where, given the current weights, we sample worlds using Gibbs sampling and compute the gradient based on the estimated expectations over $\mathcal{I}$ and $\mathcal{C}$ properties. Algorithm 1 summarizes our approach.

\begin{algorithm}[!t]{
\small
\label{alg:wvec}
\linesnumbered
\caption{Learning Potentials}
\KwIn{HMLN structure $\mathcal{H}$, Training data $\mathcal{D}$, caption $C$ generated for test image $I$}
\KwOut{Learned distribution $P(\cdot)$ with potentials $\phi_1\ldots\phi_n$}
\tcp{Normalization}
Normalize $\mathcal{H}$ w.r.t to $C$\\
\tcp{Contrastive Divergence}
$\phi_1\ldots\phi_n$ $=$ Potentials in normalized $\mathcal{H}$\\
$\mathcal{D}'$ = Training data where $\mathcal{D}'\subseteq \mathcal{D}$ corresponds to $\phi_1\ldots\phi_n$\\
${\bf w}$ = Initialize Weights for $\phi_1\ldots\phi_n$\\
\For{$t$ from 1 to $T$}
{
${\bf x}^{(1)}\ldots {\bf x}^{(K)}$ $=$ Samples from $P_{\bf w}$ using Gibbs sampling\\
\For{each weight $w_i$}
Estimate expected value in Eq.~\eqref{eq:grad} from ${\bf x}^{(1)}\ldots {\bf x}^{(K)}$\\
Compute gradient $\frac{\partial(\ell({\bf w},{\bf x}))}{\partial{w_i}}$ in Eq.~\eqref{eq:grad}\\ 
Update $w_i$ $=$ $w_i+\eta*\frac{\partial(\ell({\bf w},{\bf x}))}{\partial{w_i}}$ ($\eta$ is the learning rate)\\
}
return $\phi_1\ldots\phi_n$
}
\end{algorithm}
\normalsize

\subsection{Generating Explanations}

For models that generate samples, we can quantify {\em bias} in the generated samples using importance weighting~\cite{grover2019bias}. Specifically, suppose samples are generated under a distribution $P_{\theta}$, the bias w.r.t the true distribution $P$ is quantified with the importance weight, i.e., for a sample ${\bf x}$ the ratio $\frac{P({\bf x})}{P_\theta({\bf x})}$. 
To apply this technique in our approach, we perform importance weighting over potential functions. Specifically, let caption $C$ be generated for a test image $I$ and let ${\bf x}$ represent the ground predicates extracted from $C$. Let $\mathcal{H}'$ be the HMLN normalized for $I$. We learn the prior distribution $P(\cdot)$ over $\mathcal{H}'$ as described in the previous section. The {\em Markov blanket} of ${\bf x}$, $MB({\bf x})$ consists of all ground predicates in $\mathcal{H}'$ that are related to ${\bf x}$. Let $\phi_1$, $\ldots$ $\phi_k$ be the potentials in the Markov blanket of ${\bf x}$. Specifically, each $\phi_i$ corresponds to a potential whose scope includes at least one ground predicate in $MB({\bf x})$. Let ${\bf Y}$ represent the union of ground predicates in the scope of $\phi_1\ldots\phi_k$. Strictly, we would require $\mathcal{H}'$ to have the exact ground predicates in ${\bf x}$ to define the Markov blanket. However, this would mean $\mathcal{H}'$ should contain every possible ground predicate that can be generated by the captioning model which is infeasible in practice. Therefore, we perform a simple {\em reification} to map ${\bf x}$ to existing ground predicates in $\mathcal{H}'$. Specifically, we reify a ground predicate $G(O_1,O_2)$ to a ground predicate $G'(O'_1,O'_2)$ in $\mathcal{H}'$ iff i) $G$ $=$ $G'$ and either $O_1$ or $O_2$ matches with $O'_1$ or $O'_2$ or ii) Both $O_1,O_2$ match with $O'_1$ or $O'_2$ and the similarity of $G$ and $G'$ (measured using word embeddings) is greater than a threshold.

\noindent{\bf Virtual Evidence.} The prior distribution over $\phi_i$ is given by $P(\phi_i)$. The conditional distribution over $\phi_i$ is $P(\phi_i|\psi({\bf x}))$. Here, $\psi({\bf x})$ is {\em virtual evidence} on the generated caption. The concept of virtual evidence was first introduced by Pearl to handle uncertainty in observations. Further, this has been used in several approaches since such as semi-supervised learning~\cite{kingma2014semi} and grounded learning~\cite{parikh2015grounded} as a form of indirect supervision (e.g. label preferences). In general, the idea of virtual evidence is to introduce a potential function over the evidence that encodes its uncertainty. 
In our case, a generated caption mentions relations which is treated as evidence and therefore we condition our distribution on this evidence. However, while these are the observed evidences, our hypothesis is that these evidences cannot be treated as direct observations but as a preference of the model to mention these relations and therefore, we encode them as virtual evidences. 

We encode $\psi({\bf x})$ as univariate potentials over the ground predicates in ${\bf x}$. Specifically, this is in the factored form $\prod_i\psi(x_i)$, where $X_i\in{\bf x}$ and $\psi(x_i)$ $=$ $e^{-\log\sigma(\epsilon-d_i)}$, where $d_i$ is the cosine similarity between the CLIP embedding of the test image $I$ and the ground predicate $X_i$. We instantiate the virtual evidence, i.e., $P(\cdot|\psi({\bf x})$ to obtain the modified distribution $\hat{P}(\cdot)$ by computing the factor product of $\psi({\bf x})$ with the potentials $\phi_1\ldots\phi_k$. Specifically, the modified potentials in $\hat{P}(\cdot)$ will be as follows. For each of the ground predicates in $\psi({\bf x})$ that are in the scope of $\phi_i$, we multiply the potential values in $P$ (corresponding to rows where the predicate-in-scope is equal to 1) with the virtual evidence. An illustrative example is shown in Fig.~\ref{fig:virt}.

\noindent{\bf Bias Quantification.} Given a sample ${\bf y}$ from $\hat{P}$, we quantify {\em bias} in the generated caption by importance weighting the sample, i.e., $\mathbb{W}=\frac{P({\bf y})}{\hat{P}({\bf y})}$. Note that for the samples to be properly weighted, whenever $P(\cdot)$ is positive (non-zero) $\hat{P}(\cdot)$ must be positive. In our case, we can see that we do not set potential values to be zero in $\hat{P}(\cdot)$ since the virtual evidence is exponentially weighted. Therefore, it follows that if $P(\cdot)>0$ then $\hat{P}(\cdot)>0$ and we have properly weighted importance weights. 

The importance weights explain the influence of the virtual evidence on the prior distribution. A small weight indicates that conditioning on the evidence increased the probability of the sample compared to the prior probability and vice-versa. We can now compute expectations over functions through Monte Carlo estimates on the importance weights. Specifically, we are interested in analyzing the the $\mathcal{I}$ property within each of the potentials to explain the generated captions. Recall that the $\mathcal{I}$ property explains that the presence of one ground predicate implies the absence of the other. Let $\mathbb{I}(\mathcal{I}_i^{(j)})$ be an indicator function that has a value 1 if the $j$-th sample indicates that the $\mathcal{I}$ property in the $i$-th potential is true and 0 otherwise. The expected value of this function corresponds to the marginal density of the $\mathcal{I}$ property.



To estimate the expectation, we use Gibbs sampling to sample from $\hat{P}(\cdot)$ and compute the importance weights for each sample.
Specifically, let ${\bf y}^{(1)},{\bf y}^{(1)},\ldots{\bf y}^{(k)}$ be k samples from $\hat{P}(\cdot)$ after the MCMC chain in the Gibbs sampler has completed mixing. The importance weights, $\mathbb{W}^{(1)}\ldots\mathbb{W}^{(k)}$ are computed over the $k$ samples. Further, we use {\em clipping} to ensure that the weights do not get very large and upper bound the weights. The marginal density of the $\mathcal{I}$ property in potential $\phi_i$ is computed as follows.


\begin{equation}
\label{eq:marg1}
    \mathbb{E}_1[\mathcal{I}_i|\psi({\bf x})]=\frac{\sum_{t=1}^T\mathbb{I}(\mathcal{I}_i^{(j)})*\max(\mathbb{W}^{(j)},1)}{\sum_{j=1}^n\max(\mathbb{W}^{(j)},1)}
\end{equation}


Note that strictly speaking, we would want the samples drawn from $\hat{P}(\cdot)$ to be independent of each other. However, Gibbs sampling will generate dependent samples. To address this, we use a commonly used MCMC technique called {\em thinning}, where we only consider samples for estimating the weight after every $m$ Gibbs iterations. This reduces the dependence among samples used to compute the explanation. Since the normalization constants in $P$ and $\hat{P}$ are different, the expected values computed from the importance weights are not unbiased but {\em asymptotically unbiased}. That is as $T\rightarrow$ $\infty$, the bias reduces to 0. In this case, the expected value converges to the marginal probability of $\mathcal{I}_i$ in the distribution $P(\cdot)$.

We now generate explanations by comparing the marginal density estimated from Eq.~\eqref{eq:marg1} with the marginal densities without the importance weighting for the same samples. Specifically,

\begin{equation}
\label{eq:marg2}
    \mathbb{E}_2[\mathcal{I}_i|\psi({\bf x})]=\frac{1}{T}\sum_{j=1}^T\mathbb{I}(\mathcal{I}_i^{(j)})
\end{equation}

The expectation $\mathbb{E}_2$ converges to the marginal density in $P(\cdot|\psi({\bf x}))$. Note that a large marginal density indicates that when instantiated with the virtual evidence, it becomes more likely that one of the ground predicates in the potential cannot explain the other and therefore increases uncertainty of the generated test caption.
We compute the distance between the marginal densities obtained from Eq.~\eqref{eq:marg1} and Eq.~\ref{eq:marg2}. Specifically, in our experiments, we used the Hellinger's distance~\cite{gonzalez2013class} to compute the difference since it is a symmetric distance (as compared to the KL divergence), though the same approach can work with other distance measures as well. We interpret the Hellinger's distance based on Table~\ref{tab:intp}. Specifically, based on this interpretation, given a test instance, we explain a test instance using training samples as follows, i) most similar samples to the test instance leaves the prior invariant, ii) samples that introduce positive bias have a larger distance from the prior but also have a smaller marginal density to reduce uncertainty and iii) samples that introduce negative bias have a larger distance from the prior but also have a larger marginal density to increase uncertainty of the generated caption. Algorithm 2 summarizes our approach.




\begin{table}
    \centering
    \resizebox{0.35\textwidth}{!}{
    \begin{tabular}{|c|c|c|}
        \hline
        $P_2$ & $\delta_H(P_1,P_2)$ & Interpretation\\
        \hline\hline
         $\updownarrow$  & $\downarrow$ & $D_t$ is similar to $D_e$\\
        $\uparrow$  & $\uparrow$ & $D_e$ is less likely to explain $D_t$\\
        $\downarrow$  & $\uparrow$ & $D_t$ is more likely to explain $D_e$\\
        \hline
    \end{tabular}
    }
    \caption{\label{tab:intp} $P_1$, $P_2$ are marginal densities corresponding to the $\mathcal{I}$ property within $\phi$. $P_1$ is the importance weighted marginal density, $P_2$ is the marginal density for the evidence instantiated distribution. $\delta_H(P_1,P_2)$ is the Hellinger distance. $D_t$ is the training instance corresponding to $\phi$ and $D_e$ is the test instance.}
\end{table}

\begin{algorithm}[!t]{
\small
\label{alg:bias}
\linesnumbered
\caption{Bias Quantification}
\KwIn{$P(\cdot)$ with potentials $\phi_1\ldots\phi_n$, caption $C$ generated for test image $I$}
\KwOut{Bias quantifiers for $\phi_1\ldots\phi_n$}
\tcp{Instantiate Virtual Evidence}
${\bf x}$ $=$ Ground predicates observed in $C$\\
${\bf Y}$ $=$ Ground predicates in $MB({\bf x})$\\
\For{each $X_i\in{\bf x}$}
{
    $d_i$ $=$ similarity between CLIP embedding for $I$ and $X_i$\\
    $\psi(X_i)$ $=$ $e^{-\log\sigma(\epsilon-d_i)}$\\ 
}

$\psi({\bf x})$ $=$ $\cup_i\psi(X_i)$\\
\For{each $\phi_i$}
{
    $\phi'_i$ $=$ $\phi_i\times\psi({\bf x})$\\
}

\tcp{Importance Weighting}
$\hat{P}({\bf Y})$ $=$ Distribution with potentials $\{\phi'_i\}_{i=1}^n$\\
Burn-in Gibbs sampler for $\hat{P}({\bf Y})$\\
\For{$t$ $=$ 1 to $T$}
{
    Draw ${\bf y}^{(t)}$ $\sim$ $\hat{P}({\bf Y})$\\
    \For{each $\phi_j$}
    {
        Compute Importance weight $\mathbb{W}^{(j)}$\\
        $\overline{P}_1(\mathcal{I}_j)$ $=$ Update Monte Carlo Estimate for marginal probability of $\mathcal{I}_j$ using Eq.~\eqref{eq:marg1}\\
        $\overline{P}_2(\mathcal{I}_j)$ $=$ Update Monte Carlo Estimate for marginal probability of $\mathcal{I}_j$ using Eq.~\eqref{eq:marg2}\\
    }
}
$\delta_{H_1},\delta_{H_2},\ldots \delta_{H_n}$ $=$ Hellinger's distances between ($\overline{P}_1(\mathcal{I}_j),\overline{P}_2(\mathcal{I}_j)$)\\
Return $\delta_{H_1},\delta_{H_2},\ldots \delta_{H_n}$\\
}
\end{algorithm}
\normalsize

\section{Experiments}

The goal of our experiments is to evaluate the interpretability of explanations generated by our approach for 4 state-of-the-art captioning systems, for non-technical as well as technical users. To do this, we designed a Amazon Mechanical Turk (AMT) user study and a study among a user group with intermediate-to-expert knowledge in AI. To compare with our approach, we also designed an alternate explainer where we used an attention model for explanations. Next, we describe our experimental setup, the user study designs and the conclusions of our study\footnote{We will make additional details and results available in the supplementary material}.

\subsection{Implementation Details}

We used the following state-of-the-art approaches in our evaluation: {\tt SGAE}, {\tt AoANet}~\cite{huang2019attention}, {\tt X-LAN}~\cite{pan2020x} and {\tt M2 Transformer}~\cite{cornia2020meshed}. In each case, we used their publicly available pretrained models for the evaluation. We used the MSCOCO image captioning benchmark dataset with Karpathy's train, test, validation split~\cite{karpathy2015deep} in our evaluation method. The training data consists of 113K images whereas the validation set and test set consists of 5K and 5K images respectively. The number of captions per image is equal to 5. 
To implement our approach, we used the textual scene graph parser~\cite{schuster2015generating} to generate ground predicates from the text of captions. We used the pretrained CLIP model to obtain neural embeddings for the image and text to encode these as function values within our HMLN formulas. For weight learning in the HMLN, we set the learning rate to 0.01. For reifying the test ground predicates to approximately match them with the predicates in the training data, we used Word2Vec word embeddings. For explaining each test instance, we selected 3 examples from the training data as follows. Let $P_1,P_2\ldots P_k$ be the prior distribution and $\hat{P}_1,\hat{P}_2\ldots \hat{P}_k$ the conditional distributions corresponding to the $k$ training examples. Let $\delta_h(P_i,\hat{P}_i)$ be the Hellinger's distance between the $i$-th prior and conditional distribution. We select the following as explanations, example 1: the example with maximum positive bias ($\arg\max_i \delta_h(P_i,\hat{P}_i)+\hat{P}_2)$, example 2: the example with maximum negative bias ($\arg\max_i \delta_h(P_i,\hat{P}_i)-\hat{P}_2)$, example 3: the example with least bias ($\arg\min_i  \delta_h(P_i,\hat{P}_i)$).

\subsection{Attention-based Explanations}

Attentions have been used for explanations in language~\cite{bibal2022attention} as well as computer vision tasks~\cite{sun2020understanding}. To compare with our approach, we implemented an attention-based explanation using the weakly-supervised learning approach in~\cite{shi2020improving}. Specifically, the idea is to use the ground predicates extracted from the captions to weakly supervise a visual attention model. For this, we use the deep multiple instance learning (MIL) model in~\cite{ilse2018attention}. Corresponding to each predicate symbol (we limit to 200 of the most frequently occurring predicates), we have a MIL model that is weakly supervised with positive and negative bags of visual features which are concatenations of object pairs. Specifically, the objects in the image are identified and localized using Faster R-CNN~\cite{ren2015faster}. We use ResNet-101~\cite{he2016deep} to perform object detection in the image and the visual features are extracted from this consisting of region of interest pooling for each of the bounding boxes. The pooling function that represents a  predicate symbol combines the object-pairs in a {\em permutation-invariant} manner. To do this learning, we use the Gated-attention based pooling function in ~\cite{ilse2018attention}. Details of this model are provided in the supplementary section. To obtain explanations, for a generated test caption, we extract the ground predicates and infer from the MIL model specific to the predicate symbol the object pair in the image related by that predicate that has maximal and minimal attentions. The maximal and minimal attention objects pairs in the image serve as contrasting explanations for what a captioning model must pay attention to in generating a caption.

\subsection{AMT Study}

\begin{figure*}
    \subfigure[]{\includegraphics[scale=0.45]{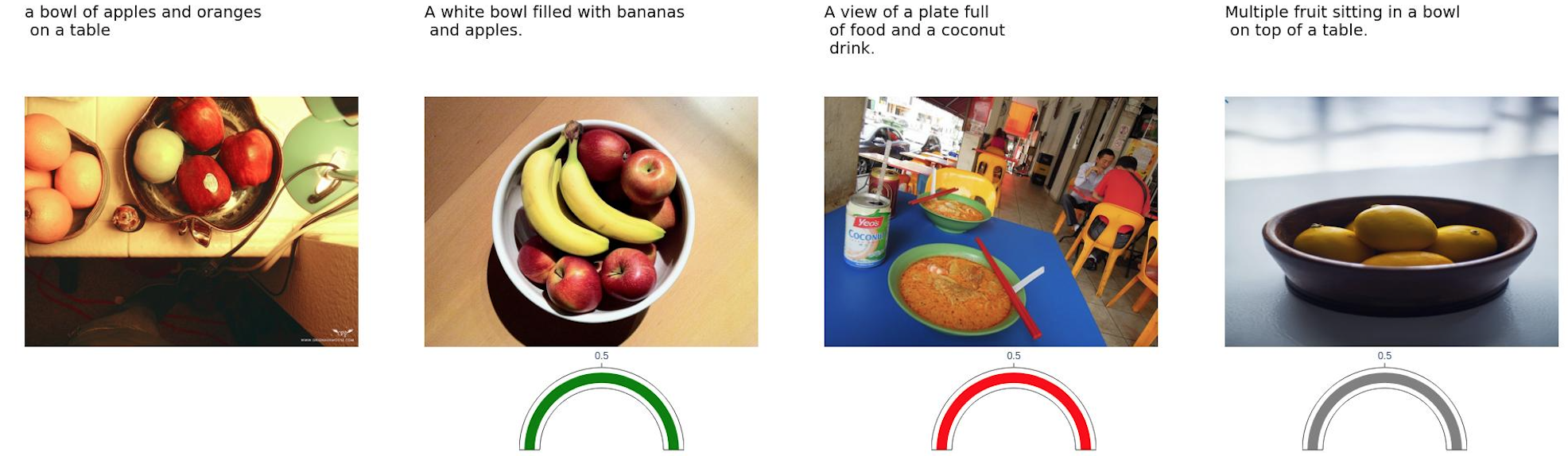}}
    \subfigure[]{\includegraphics[width=0.45\textwidth]{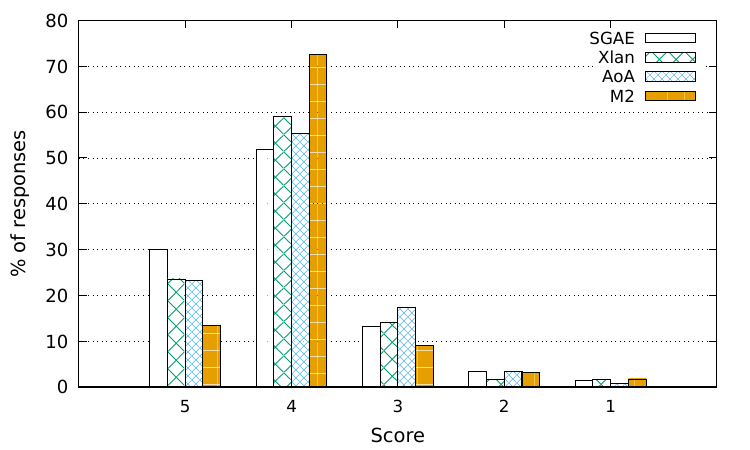}}
    \subfigure[]{\includegraphics[width=0.45\textwidth]{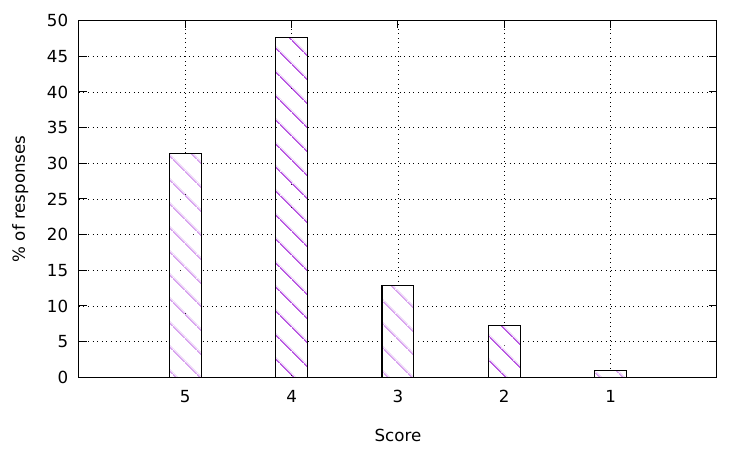}}
    \caption{\label{fig:turkresults} (a) Example explanation used in the user study (b) comparing responses from AMT users across models (higher is better) (c) Comparing technical user responses (higher is better).}
\end{figure*}

We provided users with the instructions regarding the three examples (image, caption pairs from training data) that explain the AI system's learning to generate its caption for the test image. Specifically, we had the following instructions accompanied with the image shown in Fig.~\ref{fig:turkresults} (a).
    \begin{itemize}
            \item Image 2 (\textcolor{green}{green} indicator): Related to Image 1 but also has visible differences (visually and/or in text of caption) which helps the AI extend its understanding.
            \item Image 4 (\textcolor{gray}{gray} indicator): Related to test image but with smaller differences (visually and/or in text of caption) and thus may be of limited value to the AI to learn something new.
         \item Image 3 (\textcolor{red}{red} indicator): Related test image but most likely distorts the AI's understanding of test Image due to notably larger differences (visually and/or in text of caption).
    \end{itemize}
    We then asked them to rate the explanation on a 5-point Likert scale.
    \textbf{Question:} Did the above explanation match with your interpretation of how the AI system learns to understand the test image based on the examples?

    We used a total of 1000 AMT workers to obtain responses. Each explanation was given by 5 different workers. The results on captions generated from the 4 different models are shown in Fig.~\ref{fig:turkresults} (b). As seen here, for all the systems, majority of the users were able to interpret the explanation generated by our approach (evidenced by scores of 4 or above). The inter-rater agreement is shown in Table~\ref{tab:kappa1}. As seen in this table, the raters seemed to be in agreement in most of the cases. The average scores for each model is shown in Table~\ref{tab:avgresponse}. As seen here, AoA and SGAE obtained the best responses for explanations generated from their captions. We performed a paired t-test to analyze the significance of differences in responses across different models. These results are shown in Table~\ref{tab:ttest}. As seen here, the differences were statistically significant except when we paired the responses to explanations over M2 captions with Xlan and SGAE. Thus, based on our results, AoA was the most explainable model among the four that we compared.

\begin{table}[!t]
    \centering
    \begin{tabular}{|c|c|c|c|c|c|}
        \hline
        \textbf{..} & \textbf{$R_1$} & \textbf{$R_1$} & \textbf{$R_1$} & \textbf{$R_1$} & \textbf{$R_1$}\\
        \hline
        \hline
        \textbf{$R_1$} & 1 & 0.3556 & 0.2251 & 0.3278 & 0.3230 \\ 
        \textbf{$R_1$} &  & 1 & 0.3344 & 0.3556 & 0.2857 \\
        \textbf{$R_1$} &  &  & 1 & 0.323 & 0.2045  \\        
        \textbf{$R_1$} &  &  &  & 1 & 0.192 \\ 
        \textbf{$R_1$} &  &  &  &  & 1 \\
        \hline
    \end{tabular}
    \caption{Cohen kappa scores for 5 reviewers.}
    \label{tab:kappa1}
\end{table}

    Next, we performed a study with technical users with intermediate to expert knowledge in AI. Specifically, this comprised of undergraduate seniors who had taken AI classes and Ph.D. students who work in AI. We had 25 participants in this study and here, we gave each participant 10 explanations to rate on the same 5-point Likert scale. The question and design was exactly the same as the AMT study. Here due to the small number of participants, we only used the best performing model AoA's explanations. The results of this study are shown in Fig.~\ref{fig:turkresults} (c). The results in this mirror that of our AMT study and it shows that technical users also found the explanation to be as interpretable as AMT users.

    Next, we performed an AMT study with the attention-based explanations. Specifically, here we showed to the user, the highest and lowest attention object pairs (from among the ground predicates extracted from the caption) and asked users to rate if the highest attention pair explained the AI's understanding of the image and similarly rate if the lowest attention pair explained the AI's understanding of the image. For attentions to be an effective explainer, users should be able to distinguish between the highest and lowest attention pairs. That is we would expect the higher attention object pairs to be more important in understanding the image as compared to the lower attention object pairs. However, results from the study clearly showed the opposite effect. That is, there was no significant difference between the responses we obtained for both the high and low attention object pairs. Specifically, the average scores for the high and low attention object pairs were 2.99 and 2.855 respectively with no significant difference. Thus, we can conclude that attentions did not provide significant interpretability as explanations.

\subsection{Bias Quantification}

\begin{figure*}
    \centering
   \subfigure[SAGE]{\includegraphics[scale=0.2]{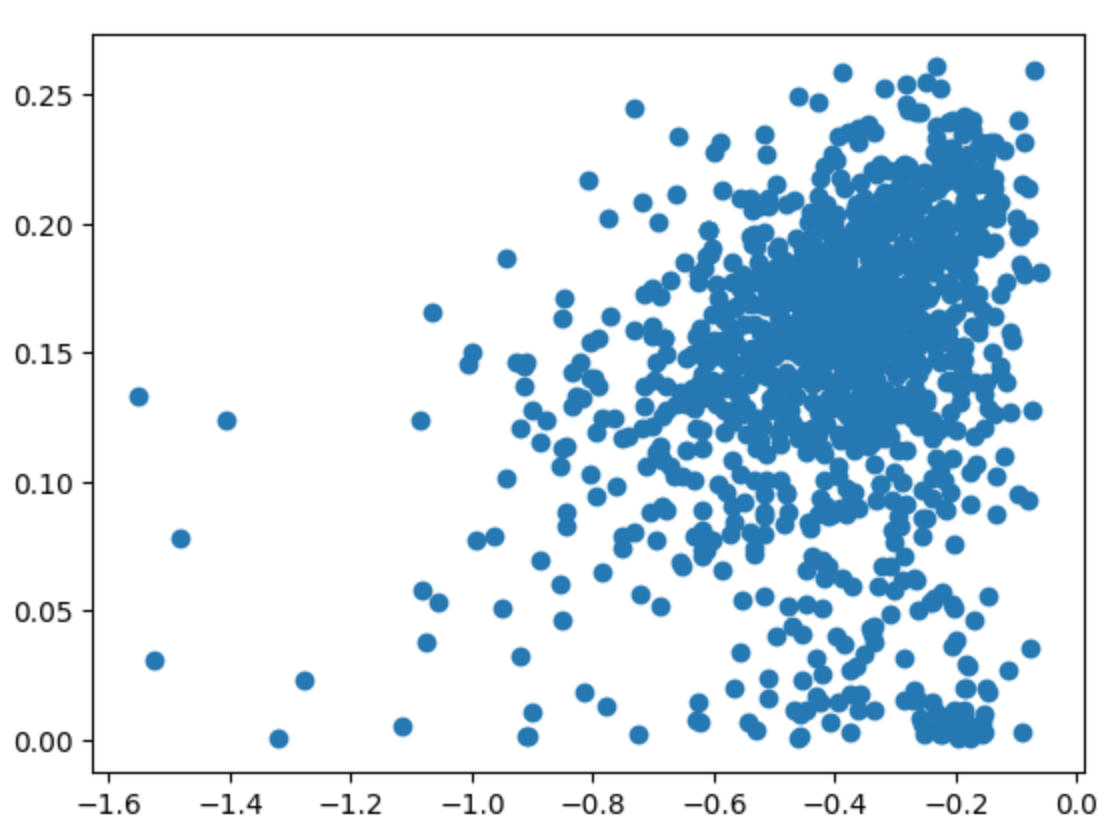}}
   \subfigure[AoA]{\includegraphics[scale=0.2]{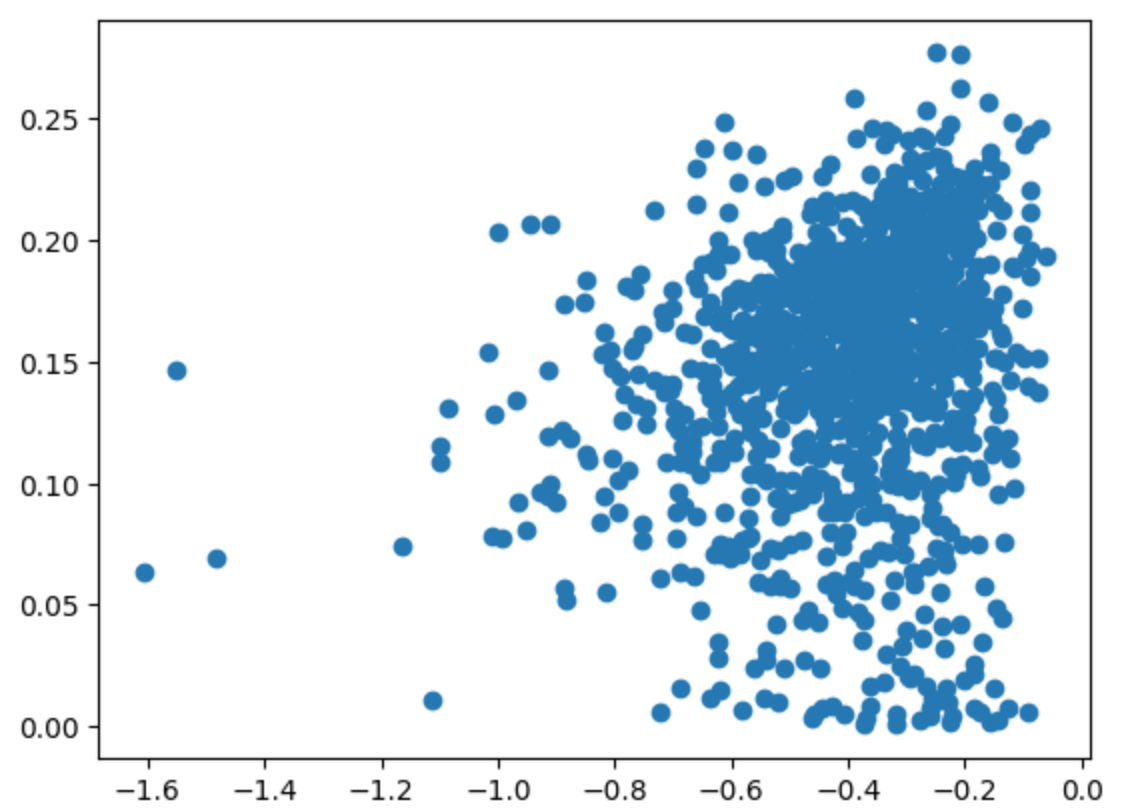}}
   \subfigure[M2]{\includegraphics[scale=0.2]{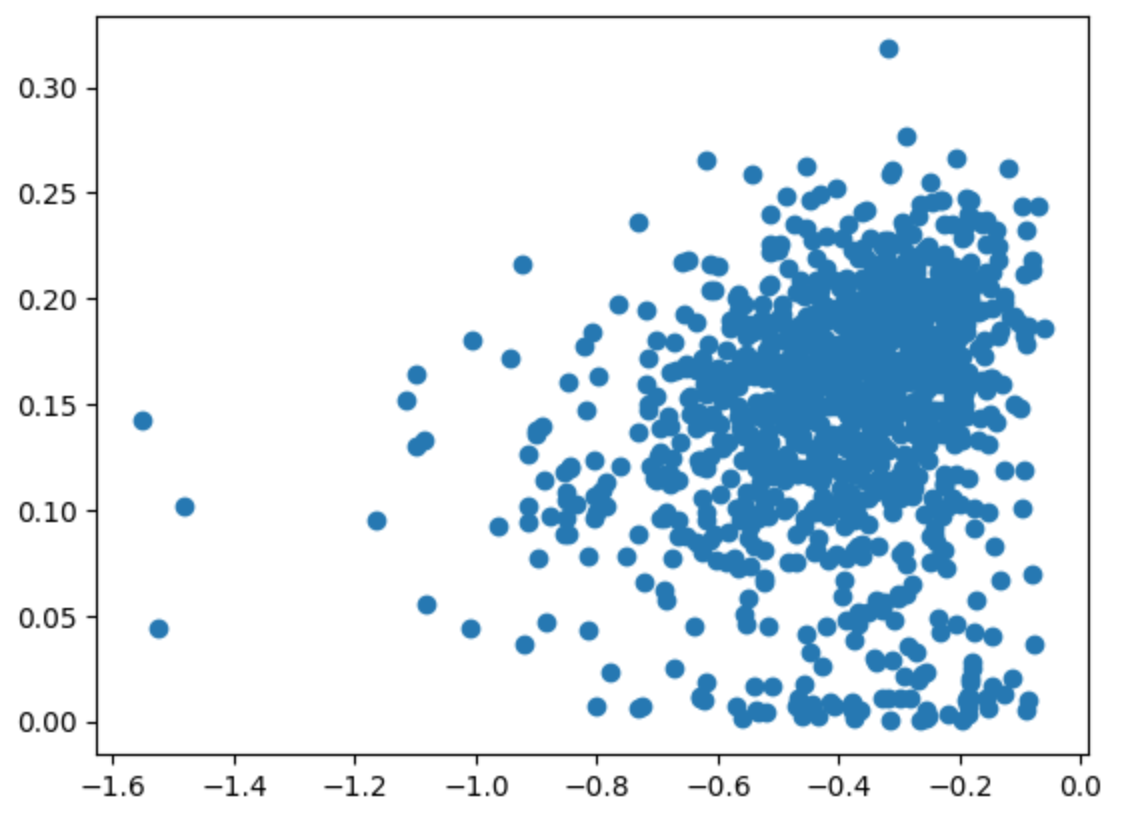}}
   \subfigure[Xlan]{\includegraphics[scale=0.2]{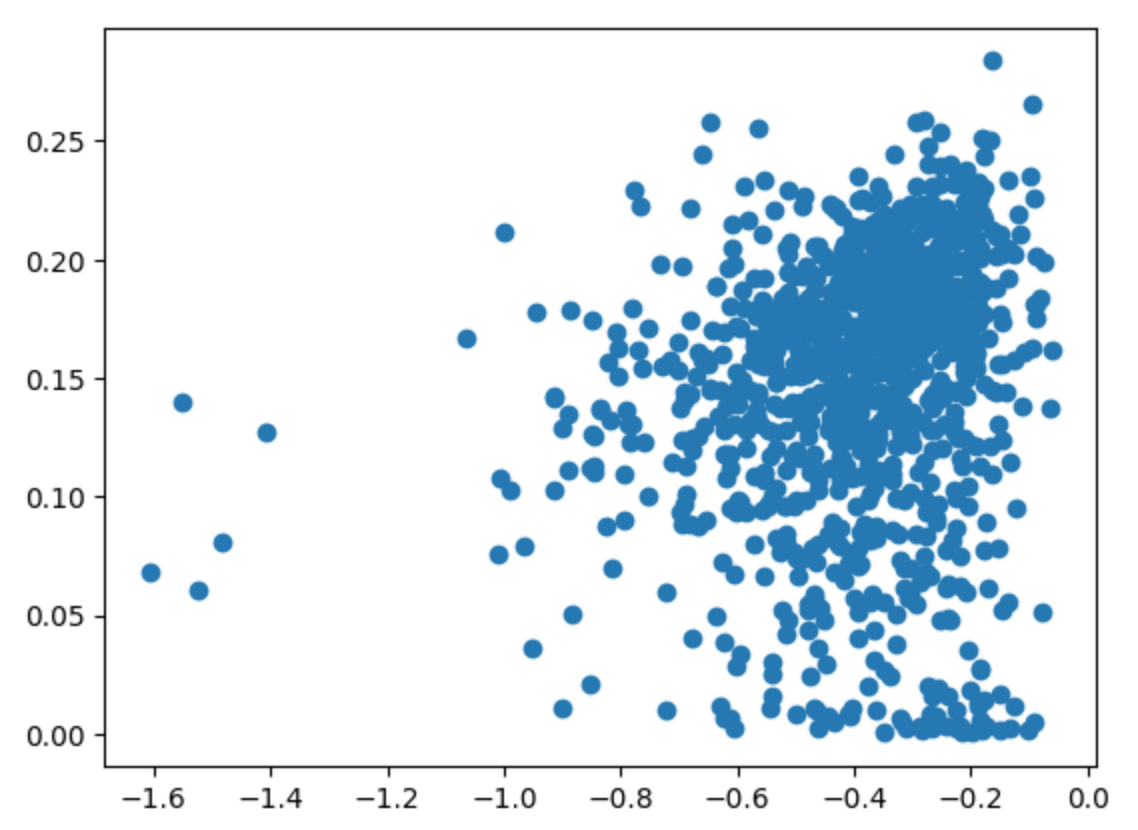}}
    \caption{The Y-axis shows the maximum Hellinger's distance between the prior and the conditionals and the X-axis shows the negative log-sigmoid of the average human-written CLIPScores (closer to 0 is a better score).}
    \label{fig:bias}
\end{figure*}

Evaluating if the distance between the prior and evidence conditioned distributions truly capture the explanation needed to generate the caption is challenging in the absence of ground truth. Therefore, we designed the following approach to evaluate our bias quantification. We used CLIPScore~\cite{hessel2021clipscore} on the ground-truth captions for the test data, i.e., captions written by humans and computed the average score over the 5 captions. We consider the maximum Hellinger's distance between prior and conditionals (considering only conditional distributions larger than a threshold of 0.75), namely, the training examples that are more useful in learning the caption, and compare the distances in these examples with the average CLIPScore. The results are shown in Fig.~\ref{fig:bias}. As seen in these results, larger average CLIPScores which are perhaps indicative that the image was explainable by humans also implies that our approach in such cases will provide an explanation with examples from more diverse contexts (shown by the larger distance between distributions).









\begin{table}
    \centering
    \begin{tabular}{|c|c|c|c|}
        \hline
        \textbf{AoA} & \textbf{M2} & \textbf{Xlan} & \textbf{SGAE}\\
        \hline
        \hline
        4.0744 & 3.94 & 3.892 & 4.017 \\ 
        \hline
    \end{tabular}
    \caption{Average response value for each model (higher is better).}
    \label{tab:avgresponse}
\end{table}

\begin{table}[!t]
    \centering
    \resizebox{0.4\textwidth}{!}{
    \begin{tabular}{|c|c|c|c|c|}
            \hline
		& \textbf{AoA} & \textbf{M2} & \textbf{Xlan} & \textbf{SGAE}\\
            \hline
            \hline
            \textbf{AoA} & & 2.087 & 0.912 & 3.008\\
            \textbf{M2} & & &\textcolor{red}{-0.78} & \textcolor{red}{-1.9}\\
            \textbf{Xlan} & & & & 1.2\\
		\hline
    \end{tabular}
    }
    \caption{\label{tab:ttest} Paired t-test results, values in red were not statistically significant ($p>0.05$).}
\end{table}

\section{Conclusion}

Explaining how AI systems achieve multimodal integration is a difficult task. In this paper, we considered visual captioning, arguably a task that has seen great progress over the last several years. However, if one needs to explain to an end-user how the AI systems learns this task, it remains challenging. In this paper, we developed a novel framework for example-based explanations in visual captioning. Specifically, we modeled a Hybrid Markov Logic Network (HMLN) to model the distribution over relations observed in the training data. We then explained a generated caption by inferring which examples are the best samples that the model could have learned from, based on observing the generated caption. We ran a comprehensive user study on Amazon Mechanical Turk to show that explanations generated by our model are highly interpretable. In future, we will build upon this framework to explain other complex generative models such as Visual Question Answering.

\bibliographystyle{ACM-Reference-Format}
\bibliography{ref}

\end{document}